\documentclass[letterpaper, 10 pt, conference]{ieeeconf}  

\IEEEoverridecommandlockouts                              

\usepackage{cite}
\usepackage{float}
\usepackage{amsmath,amssymb,amsfonts}
\usepackage{algorithmic}
\usepackage{graphicx}
\usepackage{textcomp}
\usepackage{xcolor}
\def\BibTeX{{\rm B\kern-.05em{\sc i\kern-.025em b}\kern-.08em
    T\kern-.1667em\lower.7ex\hbox{E}\kern-.125emX}}

\usepackage{booktabs}
\usepackage{bbding}
\usepackage{pifont}
\usepackage{wasysym}
\usepackage{amssymb}
\usepackage{makecell}
\usepackage{multirow}
\usepackage{rotating}
\usepackage{siunitx}
\usepackage{hyperref}
\usepackage{cleveref}

\usepackage{pifont}
\usepackage[nohyperlinks, printonlyused, withpage, smaller]{acronym}
\usepackage{amssymb}

\usepackage{footnote}
\usepackage{fancyhdr}
\usepackage[absolute]{textpos}
\acrodef{CTRA}[CTRA]{constant turn rate and acceleration}
\acrodef{VRU}[VRU]{vulnerable road user}
\acrodef{DSC}[DSC3D]{DeepScenario Open 3D Dataset}
\acrodef{SfM}{Structure-from-Motion}
\acrodef{MVS}{Multi-View Stereo}
\acrodef{GPS}{Global Positioning System}
\acrodef{BA}{Bundle Adjustment}
\acrodef{UTM}{Universal Transverse Mercator}
\acrodef{SIFT}{Scale-Invariant Feature Transform}
\acrodef{NURBS}[NURBS]{Non-Uniform Rational B-Splines}
\acrodef{DSC-SIFI}[SIFI]{Fabulous Sindelfingen}
\acrodef{DSC-STR}[STR]{Stunning Stuttgart}
\acrodef{DSC-MUC}[MUC]{Great Munich}
\acrodef{DSC-BER}[BER]{Vibrant Berlin}
\acrodef{DSC-SFO}[SFO]{Visionary San Francisco}
\def\our_name{\acs{DSC}}
\acrodef{HD}[HD]{High-Definition}
\acrodef{PnP}[PnP]{Perspective-n-Point}
\acrodef{RMSE}[RMSE]{Root Mean Squared Error}
\acrodef{ADE}[ADE]{Average Displacement Error}
\acrodef{FDE}[FDE]{Final Displacement Error}
\acrodef{TTC}[TTC]{Time-to-Collision}
\acrodef{PET}[PET]{Post-Encroachment Time}
\acrodef{GCP}[GCP]{Ground Control Point}

\newcommand{\mycomment}[1]{}
\newcommand*{\inparagraph}[1]{\noindent\textbf{#1}\hspace{0.5em}}

\begin{document}

\title{Highly Accurate and Diverse Traffic Data:\\The DeepScenario Open 3D Dataset}

\author{
Oussema Dhaouadi$^{1,2,3,^*}$,
Johannes Meier$^{1,2,3,^*,\dagger}$,
Luca Wahl$^{1}$,
Jacques Kaiser$^{1}$,
Luca Scalerandi$^{1}$,\\
Nick Wandelburg$^{1}$, 
Zhuolun Zhou$^{1}$,
Nijanthan Berinpanathan$^{1}$,
Holger Banzhaf$^{1}$,
Daniel Cremers$^{2,3}$
\\ \\
$^{1}$ DeepScenario\quad $^{2}$ TU Munich\quad $^{3}$ Munich Center for Machine Learning\\ \\
 {\small Project Page:} {\footnotesize \textcolor[HTML]{F1238F}{\textbf{\url{https://deepscenario.github.io/DSC3D/}}}}
}



 \begin{textblock}{175}(16,263)   
 \scriptsize
 \setlength{\fboxsep}{3pt}
 \framebox{\parbox{\textwidth}{
 Accepted at IEEE Intelligent Vehicles Symposium (IV) 2025. \copyright 2025 IEEE.  Personal use of this material is permitted.  Permission from IEEE must be obtained for all other uses, in any current or future media, including reprinting/republishing this material for advertising or promotional purposes, creating new collective works, for resale or redistribution to servers or lists, or reuse of any copyrighted component of this work in other works.
 }}
 \end{textblock}

\twocolumn[{%
\renewcommand\twocolumn[1][]{#1}%
\maketitle
\begin{center}
\vspace{-0.5cm}
\includegraphics[width=\linewidth]{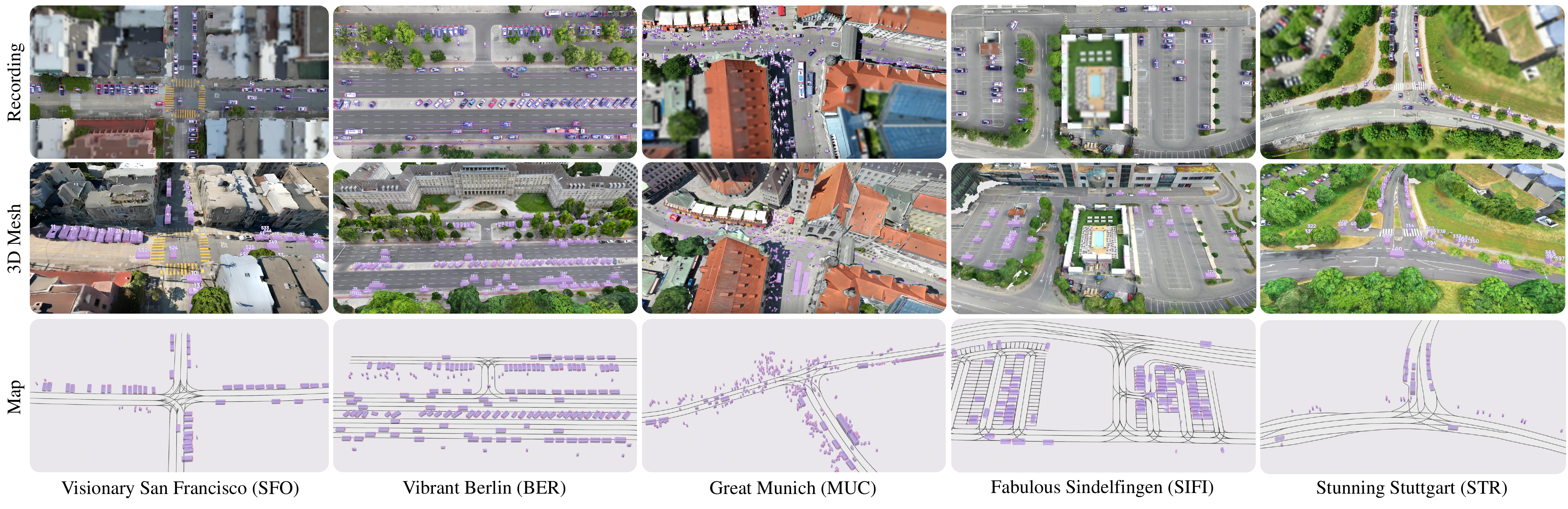}
\footnotesize{Fig. 1. Visualization of traffic participants across five locations in our \ac{DSC}: Top row shows 3D bounding boxes in recordings, middle row in the reconstructed mesh, and bottom row in the map view.}
\end{center}%
}]
\setcounter{footnote}{0}
\renewcommand{\thefootnote}{}
\footnotetext{*Shared first authorship, $\dagger$ Corresponding author.}
\footnotetext{TUM: {\tt \{oussema.dhaouadi, j.meier, cremers\}@tum.de}}
\footnotetext{DeepScenario: \tt \{lucaw,firstname\}@deepscenario.com}
\footnotetext{This work is a result of the joint research project STADT:up. The project is supported by the German Federal Ministry for Economic Affairs and Climate Action (BMWK), based on a decision of the German Bundestag. The author is solely responsible for the content of this publication. This work was also supported by the ERC Advanced Grant SIMULACRON.}
\renewcommand{\thefootnote}{\arabic{footnote}} 

\setcounter{figure}{1}

\begin{abstract}
Accurate 3D trajectory data is crucial for advancing autonomous driving. Yet, traditional datasets are usually captured by fixed sensors mounted on a car and are susceptible to occlusion. Additionally, such an approach can precisely reconstruct the dynamic environment in the close vicinity of the measurement vehicle only, while neglecting objects that are further away. In this paper, we introduce the \acf{DSC}, a high-quality, occlusion-free dataset of 6 degrees of freedom bounding box trajectories acquired through a novel monocular camera drone tracking pipeline. Our dataset includes more than 175,000 trajectories of 14 types of traffic participants and significantly exceeds existing datasets in terms of diversity and scale, containing many unprecedented scenarios such as complex vehicle-pedestrian interaction on highly populated urban streets and comprehensive parking maneuvers from entry to exit. \acs{DSC} dataset was captured in five various locations in Europe and the United States and include: a parking lot, a crowded inner-city, a steep urban intersection, a federal highway, and a suburban intersection. Our 3D trajectory dataset aims to enhance autonomous driving systems by providing detailed environmental 3D representations, which could lead to improved obstacle interactions and safety. We demonstrate its utility across multiple applications including motion prediction, motion planning, scenario mining, and generative reactive traffic agents. Our interactive online visualization platform and the complete dataset are publicly available at \href{https://app.deepscenario.com}{app.deepscenario.com}, facilitating research in motion prediction, behavior modeling, and safety validation.
\end{abstract}

\section{Introduction}

Development and validation of autonomous driving systems is among those extremely challenging technological undertakings of our time. After the promising technical demonstrations, the complexity of real-world traffic scenarios still raises numerous challenges, especially in urban environments, where comprehensive datasets are critical in capturing traffic dynamics \cite{inD,rounD}.

Traditional data collection methods using vehicle-mounted sensors or stationary infrastructure cameras are constrained in capturing the complexities of traffic interactions. Those systems suffer from occlusion and a limited field of view \cite{openDD,CitySim}. Current datasets usually have limited ability to capture a wide spectrum of vehicle interactions with \acp{VRU} - most notably in dense urban settings where comprehensive scene understanding is most needed.

A critical, yet often neglected, aspect is related to the three-dimensional information in driving scenarios. Driving decisions under real traffic are critically depending on the 3D spatial relations between road users, infrastructure, and terrain geometry, especially in complex urban scenarios. While drones have emerged as an enabling solution to collect traffic data, current drone-based datasets typically provide 2D or 2.5D representations and focus on limited geographical locations or specific scenarios~\cite{inD,highD,rounD}.

To overcome the aforementioned limitations, we present the \acf{DSC}: a large-scale drone trajectory dataset, comprehensively capturing a rich 3D scene of diverse locations across both, Germany and the United States, with five locations, as shown in Figure 1. It features:

\begin{itemize}
    \item \ac{DSC-MUC}: Represented by an area with high pedestrian density in downtown Munich. In this environment, vehicles cautiously move amidst the heavy crowd with cyclists in every direction.
    \item \ac{DSC-BER}: A federal highway (German: Bundesstraße) with a speed limit of 50 km/h. This section is representative of the traffic dynamics in large arterial roads within the urban context.
    \item \ac{DSC-SIFI}: Interesting parking maneuvers that provide insight into how vehicles position and navigate around busy environments.
    \item \ac{DSC-STR}: This is a T-intersection with no traffic lights. Unique behaviors are emphasized in the interaction of vehicles at this spot, using manual control, where there is no automated management of traffic. 
    \item \ac{DSC-SFO}: This intersection has steep roads and is unsignalized, providing challenging driving scenarios that will go well for the study of vehicle dynamics in hilly and complex environments.
\end{itemize}

This diversity across locations enables the development of more robust and generalizable autonomous systems, allowing for the exploration of various traffic patterns.

\ac{DSC} distinguishes itself through several key features, including the most extensive classification system among existing datasets (14 categories), full 3D bounding boxes for all tracked objects, and detailed 3D mesh representations with geo-referenced \ac{HD} maps in OpenDRIVE~\cite{opendrive} format.

Our main contributions are:
\begin{itemize}
\item A high-precision 3D tracking pipeline achieving median positional errors of 4.8 centimeters
\item A comprehensive urban dataset of 15 hours and 175K unique trajectories with 14 categories. 
\item An interactive web interface for download, visualization, and with additional information 
\item Diverse scenarios including challenging conditions such as steep road grades (up to 20\%), complex intersections, high-density urban traffic, federal highway scenarios, and intricate parking maneuvers
\item Extensive analysis demonstrating utility across multiple autonomous driving tasks
\end{itemize}

\section{Related Work}
Datasets for planning and prediction tasks can generally be divided into perception-based and trajectory-based datasets~\cite{exID}.

Examples of perception datasets include KITTI~\cite{KITTI}, Waymo~\cite{Waymo}, NuScenes~\cite{nuScenes}, Lyft \cite{lyft}, and Argoverse 2~\cite{Argoverse2}. All these aforementioned provide manually labeled 3D bounding boxes with image data. Though essential for perception algorithms, these datasets have lower frame rates~\cite{KITTI, nuScenes}, fewer trajectories, and occlusion issues due to the egocentric viewpoint of data acquisition.

Trajectory datasets, which are mostly auto-labeled, provide larger-scale annotations. The ego-vehicle datasets, such as WOMD-LiDAR~\cite{WOMD} and IAMCV~\cite{IAMCV}, capture diverse scenarios but are still prone to occlusions. The infrastructure camera datasets, like NGSIM~\cite{NGSIM}, BIWI~\cite{BIWI}, have fewer occlusions but suffer from low resolution~\cite{NGSIM} or lack of class diversity~\cite{BIWI}. Cooperative datasets, like V2AIX~\cite{V2AIX}, V2X-Seq~\cite{V2X-Seq}, MONA~\cite{MONA}, improve the accuracy due to multi-perspective recordings but are limited by geographic diversity~\cite{MONA,V2AIX} or international availability~\cite{V2X-Seq}. 

In contrast, drone-based datasets provide enhanced visibility and high-quality trajectory capture through their elevated perspective. They offer near-complete coverage of the scene while only facing limitations with building occlusions and low-height details, thus complementing traditional ground-based collection methods.

\inparagraph{University Campus Drone Datasets}
University Campus datasets, such as the Stanford Drone dataset~\cite{Stanford_Drone}, CITR/DUT, and CTV~\cite{CTV}, focus on interactions between pedestrians and cyclists in controlled experiments. Both, CITR/DUT~\cite{CITR_DUT} and CTV~\cite{CITR_DUT}, are annotated based on data from staged scenarios. Thus, these two datasets have limited generalization for real world applications. The Stanford Drone dataset~\cite{Stanford_Drone} and CTVV~\cite{CTV} provide trajectories in pixel space rather than in metric space. Besides that, the resolution of the Stanford Drone dataset is relatively low at $595 \times 326$ pixels and has some inaccuracies with respect to its trajectory data~\cite{CTV}.

\inparagraph{Highway Drone Datasets}
All highway datasets, such as DORA~\cite{JKU_DORA}, highD~\cite{highD}, exiD~\cite{exID}, Automatum~\cite{Automatum}, RWTH Aachen~\cite{rwth_aachen}, and AD4CHE~\cite{AD4CHE} are high-speed environments that capture extensive trajectories of vehicles. However, these lack critical orientation information~\cite{highD,rwth_aachen,JKU_DORA} and do not include \acp{VRU}. Furthermore, DORA~\cite{JKU_DORA} does not contain the dimensions of detected objects which restricts their usefulness in scenario mining. All of these datasets have their merit w.r.t specific applications but fall quite short when it comes to capturing the true complexity of an urban driving scenario.

\inparagraph{Urban Drone Datasets}
Urban datasets like rounD~\cite{rounD}, OpenDD~\cite{openDD}, inD~\cite{inD}, and SIND~\cite{SIND} capture richer interactions at roundabouts and intersections. However, they underrepresent \ac{VRU}s (e.g., OpenDD~\cite{openDD} contains 81,372 vehicle but only 3,402 pedestrian trajectories).

\inparagraph{Diverse Drone Datasets}
Very few datasets span multiple scene types. CitySim~\cite{CitySim} contains highways, roundabouts, and intersections, but the data is no longer publicly available and object dimensions are not included. The Interaction dataset~\cite{INTERACTION} is geographically diverse but only contains two classes and lacks important information such as pedestrian dimensions and orientation. 

Our dataset overcomes these limitations in the following key advancements: It is the first publicly available dataset that contains inner-city environments and parking areas while providing 3D metric annotations with 6DoF poses and 3D object dimensions, thus enabling the analysis of non-planar traffic scenarios, allowing for more robust algorithm development for planning and prediction tasks.

\section{Method}
\begin{figure*}[ht]
    \centering
    \includegraphics[width=\textwidth]{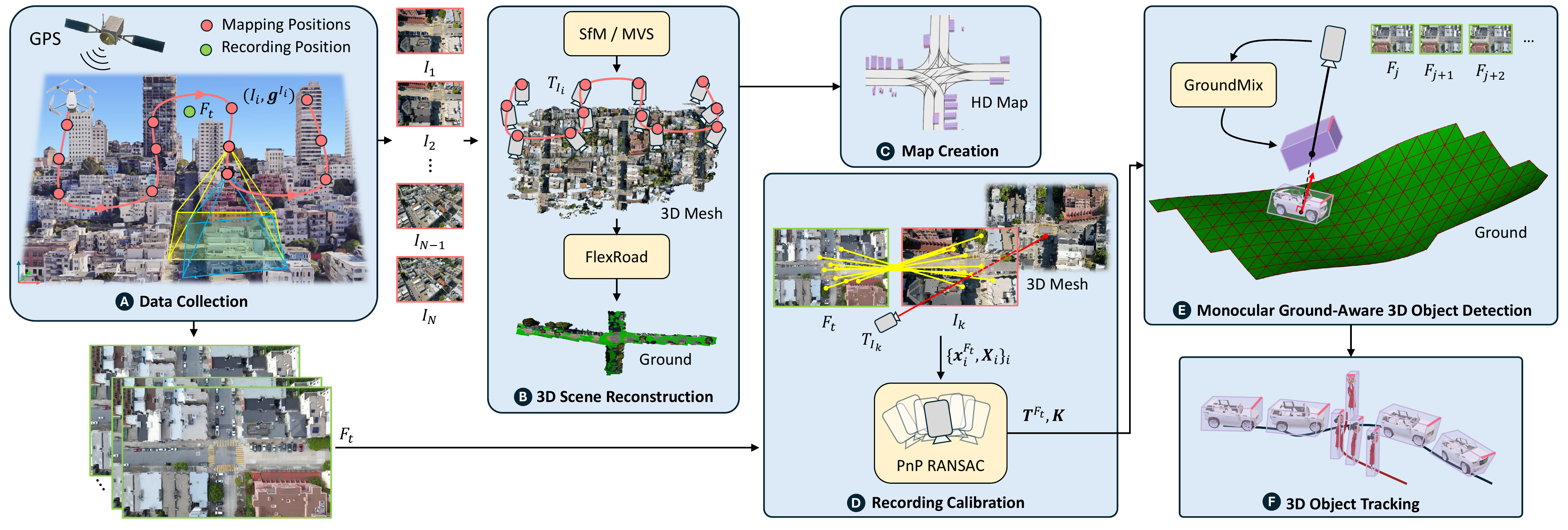}
    \caption{Our pipeline includes (A) data collection, (B) geo-referenced 3D scene reconstruction and ground generation, (C) map creation, (D) recording calibration, (E) monocular 3D object detection and detection refinement, and (F) object tracking.}
    \vspace{-15px}
    \label{fig:pipeline}
\end{figure*}

Our pipeline, as depicted in \Cref{fig:pipeline}, is divided into several steps to accurately and coherently track 3D objects. First, the pipeline processes data from monocular cameras, as described in \Cref{sec:data_collection}. Next, we calibrate all the frames of the recording by performing a geo-referenced 3D reconstruction of the environment to improve the accuracy of 3D detection, as explained in \Cref{sec:reconstruction}. The ground is refined at this step to ensure smooth road surfaces, which are essential for generating precise trajectories. We then create an \ac{HD} map, which can be used as a basis for higher-level applications such as digital twin integration and simulation (see \Cref{sec:map}). Camera calibration and localization are performed to align the global representation of bounding boxes across frames, as detailed in \Cref{sec:calibration}. Subsequently, we detect 3D object bounding boxes and enhance these detections using the smoothed ground surfaces to achieve high accuracy, as described in \Cref{sec:detection}. Finally, the pipeline performs object tracking to obtain continuous trajectories with unique IDs, as explained in \Cref{sec:tracking}.

\subsection{Data Collection}
\label{sec:data_collection}

Our approach involves the collection of data by flying commercially available DJI drones over various traffic scenarios across five selected locations with unique traffic and environmental characteristics. The recordings were made at 25Hz with downward-tilted cameras.
We perform a two-step flying of drones to capture the scenes-first, flying for mapping and capturing GPS-stamped images ${(I_i, \mathbf{g}^{I_i})}_{i=1}^N$, such that the captured areas are fully covered with enough overlap ratio. Then, recording the scenes from fixed positions by the drone for a number of hours gives the frames ${(F_t, \mathbf{g}^{F_t})}_t$.

\subsection{3D Scene Reconstruction}
\label{sec:reconstruction}
We reconstruct the complete scene, then generate an accurate and smooth ground mesh.
\subsubsection{Full Scene Reconstruction}
From $N$ sparse images $I_i$ where $1 \leq i \leq N$ and corresponding \ac{GPS} positions $\mathbf{g}^{I_i}$, we extract features and construct a pose graph. Using \ac{SfM}, we compute initial camera poses $\mathbf{T}^{I_i}_{\text{init}} = \begin{bmatrix} \mathbf{R}^{I_i}_{\text{init}} & \mathbf{t}^{I_i}_{\text{init}} \end{bmatrix}$ and a sparse point cloud. \ac{MVS} densification produces and refined poses $\mathbf{T}^{I_i}$ through \ac{BA}.

We generate a triangular mesh with vertices $\mathcal{V}$ and faces $\mathcal{F}$ and convert the \ac{GPS} positions to local \ac{UTM} coordinates to avoid numerical instabilities. To obtain the geo-referenced poses $\mathbf{T}^{I_i}$, a joint \ac{BA} minimizes both reprojection error and GPS alignment:
\begin{equation}
    \mathcal{L} = \mathcal{L}_{\text{reproj}} + \lambda \sum_i \|\mathbf{c}^{I_i} - \mathbf{g}_{\text{local}}^{I_i}\|^2,
\end{equation}
where $\mathbf{c}^{I_i} = -\mathbf{R}^{I_i\top} \mathbf{t}^{I_i}$ is the camera center and $\lambda$ balances GPS constraint with reprojection error. The mesh is then texturized and compressed while preserving geometric details.

\subsubsection{Ground Generation}
We employ FlexRoad~\cite{flexroad} in order to produce a smooth and accurate ground mesh from the full 3D scene mesh. We render an orthophoto from the textured mesh, classify the road mesh using semantic segmentation, then perform dense point sampling and filtering to separate the points which belong to the road and remove false positives, such as points from building facades. Finally, a \ac{NURBS}~\cite{nurbs_book_1997} surface is fitted to the road point cloud, optimized for an appropriate balance between accuracy and smoothness. More information can be found in~\cite{flexroad}.

\subsection{Map Creation}
\label{sec:map}
Our road network modeling with OpenDRIVE~\cite{opendrive} produces a 3D \ac{HD} map that features elevation profiles which match the actual ground surface. We optimize the road network to achieve continuity at junctions and intersections for realistic modeling.

\subsection{Recording Calibration}
\label{sec:calibration}
While drones can maintain fixed positions with gimbal stabilization, accurately referencing the camera position within the static 3D scene remains a key challenge. To ensure high-quality data, we perform precise camera localization for all frames in each recording sequence. Similar to~\cite{pollok2016visual}, we register individual frames to the reconstructed scene by matching recording frames $F_t$ with mapping images $I_k$ using state-of-the-art feature matching algorithms, such as LoFTR~\cite{loftr} and LightGlue~\cite{lightglue}.
Specifically, for frame $F_t$ with 2D points $\{\mathbf{x}^{F_t}_i\}_i$ and their corresponding 3D points~$\{\mathbf{X}_i\}_i$, we solve:

\begin{equation}
    \mathbf{T}^{F_t,*}, \mathbf{K}^{*} = \arg\min_{\mathbf{T}^{F_t}, \mathbf{K}} \sum_i \rho\left(\pi(\mathbf{K}, \mathbf{T}, \mathbf{X}_i) - \mathbf{x}^{F_t}_i\right)
\end{equation}

where $\mathbf{T}^{F_t}$ represents the frame's extrinsic parameters, $\mathbf{K}$ the camera intrinsics, $\pi$ the projection function, and $\rho$ a robust cost function within a RANSAC~\cite{ransac} framework. The resulting camera poses are further refined using Kalman filtering to ensure temporal consistency across the sequence.

\subsection{Monocular Ground-Aware 3D Object Detection}
\label{sec:detection}
Our object detection process is initialized by an estimation of 3D object positions that is further refined in subsequent processing.
\subsubsection{3D Object Detection} 
We extract 3D object information using a monocular detector based on GroundMix~\cite{GroundMix}. This method differs from existing approaches~\cite{rwth_aachen,SIND,INTERACTION,inD,CitySim}, as it conducts direct complete 3D property predictions in one step and makes use of enhanced augmentation techniques. In this paper, for each frame $F_t$, the detector predicts 2D bounding boxes, object categories, and 3D parameters such as dimensions $[l, w, h]$, 3D orientation, depth $Z_c$, and projected ground center $\mathbf{x}_p$. For readability, we omit the frame index $F_t$ from the variables of the objects appearing in the t-th frame. The 3D ground center is computed as:
\begin{equation}
\mathbf{X}_c = \mathbf{K}^{-1} \hat{\mathbf{x}}_p \cdot Z_c \, ,
\end{equation}
where $\hat{\mathbf{x}}_p$ is homogeneous pixel coordinates.

\subsubsection{Detection-Refinement}
To improve monocular depth prediction accuracy, we refine detections in multiple steps. We first refine the 3D location by computing the intersection $\mathbf{X}^{*}_c$ of a ray from the camera center through the projected 3D ground center with the ground mesh. For orientation refinement, we decompose the predicted orientation matrix into intrinsic $XYZ$ Euler angles ($\phi, \theta, \psi$):
\begin{equation}
\mathbf{R}_c = \mathbf{R}_Z(\psi) \mathbf{R}_Y(\theta) \mathbf{R}_X(\phi)\, .
\end{equation}
Following ~\cite{GroundMix}, we combine the object's ground orientation $\mathbf{R}_Z(\psi)$ with ground-normal-derived orientations $\mathbf{R}_X^{*}(\theta), \mathbf{R}_Y^{*}(\phi)$ to get ground-consistent orientation: 
\begin{equation}
\mathbf{R}^{*}_c = \mathbf{R}_Z(\psi) \mathbf{R}_Y^{*}(\omega) \mathbf{R}_X^{*}(\phi)  \, .
\end{equation}
Finally, the refined positions and orientations are transformed into world coordinates using the extrinsics $\mathbf{T}^{F_t} = \begin{bmatrix} \mathbf{R}^{F_t} & \mathbf{t}^{F_t} \end{bmatrix}$:
\begin{align}
\mathbf{X}_{w} &= \mathbf{R}^{F_t} \mathbf{X}^{*}_c + \mathbf{t}^{F_t} \, ,\\
\mathbf{R}_w &= \mathbf{R}^{F_t} \mathbf{R}^{*}_c \, .
\end{align}
\subsection{3D Object Tracking}
\label{sec:tracking}
We perform 3D tracking using a Kalman filter by calculating velocities from 3D position displacements between matched objects. Track smoothness is further enhanced using an RTS-Smoother~\cite{exID}. Further, we finetune our 3D detection model with highly curated, manually-annotated images to make sure the quality of the data is high.

\setlength{\tabcolsep}{3pt}
\begin{table*}[!ht]
\centering
\vspace{5px}
\caption{
\textbf{Drone trajectory datasets comparison: \ac{DSC} is the first to provide complete 3D annotations, showcasing high diversity, full scene information, and the most class-diversity.} \newline
\textbf{Scene Types}: \textit{I} = intersection, \textit{U} = university campus, \textit{H} = highway, \textit{H$^*$} = federal highway, \textit{R} = roundabout, \textit{P} = parking, \textit{C} = inner city. 
\textbf{Loc}: Locations; \textbf{Traj}: Numbers of Trajectories; \textbf{Pos}: Position format (px = Pixel, 2D, 3D); \textbf{Orient}: Object heading (1 DoF = single angle, 3 DoF = $SO(3)$); \textbf{Dim}: Dimensions (l, w, h in metric); \textbf{Cls}: Numbers of Categories; \textbf{Geo}: UTM-georeferenced positions. \textbf{(·)} indicates unavailability for cyclists and pedestrians.
}\vspace{-0.5em}

\begin{tabular}{c|lc|llc|ccc|rrcccccc}
\toprule

& \multirow{2}{*}{\textbf{Dataset}} & \multirow{2}{*}{\textbf{\makecell{Release \\ \hspace{-0.4em}Year}}} & \multicolumn{3}{c|}{\textbf{Diversity}} & \multicolumn{3}{c|}{\textbf{3D Scene}} & \multicolumn{7}{c}{\textbf{Annotations}}\\

& & & \textbf{Countries} & \textbf{Scenes} & \textbf{Loc} & \textbf{Map} & \textbf{Mesh} & \textbf{Video} & \textbf{Length}& \textbf{Traj} & \textbf{Pos} & \textbf{Orient} & \textbf{Dim} & \textbf{Cls} & \textbf{Geo} \\ \midrule


\multirow{13}{*}{\rotatebox{90}{\textbf{Specalized}}} & Stanford Drone \cite{Stanford_Drone}  & 2016 & USA & U & 6 & - & - & \checked & 9 h & 10k & px & - & - & 6 & - \\ 
& CITR, DUT \cite{CITR_DUT}  & 2019 &  USA & U & 2 & - & - & \checked & 0.4 h & 2 k & 2D & - & - & 3 & - \\ 
& CTV \cite{CTV} & 2023 &  Germany & U & 2 & - & - & \checked & 1.7 h & 4 k & px & 1 DoF & - & 3 & - \\ 
\addlinespace[3pt]
\cline{2-16}
\addlinespace[3pt]
& roundD \cite{rounD}  & 2020 &  Germany & R & 8 & \checked & - & - & 6 h & 14 k & 2D & 1 DoF & $(l, w)$ & 8 & \checked \\ 
& OpenDD \cite{openDD}  & 2020 &  Germany & R & 7 & \checked & - & - & 62 h & 85 k & 2D & 1 DoF & $l, w$ & 8 & \checked \\ 
\addlinespace[3pt]
\cline{2-16}
\addlinespace[3pt]
& inD \cite{inD}  & 2020 &  Germany & I & 4& \checked & - & - & 10 h & 12 k & 2D & 1 DoF & $(l, w)$ & 4 & \checked \\ 
& SIND\cite{SIND}  & 2022 &  China & I & 1 & \checked & - & - & 7 h & 13 k & 2D & (1 DoF) & $(l, w)$ & 6 & -\\ 
\addlinespace[3pt]
\cline{2-16}
\addlinespace[3pt]
& highD \cite{highD}  & 2018 &  Germany & H & 6 & \checked & - & - & 17 h & 111 k & 2D & - & $l, w$ & 2 & - \\ 
& Automatum \cite{Automatum}  & 2021 &  Germany & H & 12 & \checked & - & - & 30 h & 60 k & 2D & 1 DoF & $l, w$ & 3 & - \\ 
& exiD \cite{exID}  & 2022 &  Germany & H & 7 & \checked & - & - & 16 h & 69 k & 2D & 1 DoF & $l, w$ & 6 & \checked \\ 
& DORA \cite{JKU_DORA}  & 2022 &  Austria, Italy & H & 13 & - & - & - & 2 h & 5 k & 2D & - & - & 5 & - \\ 
& AD4CHE \cite{AD4CHE}  & 2023 &  China & H & 4 
 & \checked & - & - & 5 h & 54 k & 2D & 1 DoF & $l, w$ & 3 & - \\
& RWTH Aachen \cite{rwth_aachen}  & 2024 &  Germany & H & 1 & - & - & - & 1.5 h & 9 k & 2D & - & $l, w$ & 5 & \checked \\ 
\midrule

\multirow{3}{*}{\rotatebox{90}{\textbf{Diverse}}} 
& INTERACTION \cite{INTERACTION}  & 2019 & \makecell{China, Bulgaria, \\ \hspace{-0.4em}USA, Germany} &  H, I, R & 5 & \checked & - & - & 17 h & 40 k & 2D & (1 DoF) & ($l, w$) & 2 & - \\ 
& CitySim \cite{CitySim}  & 2024 &  USA & H, I, R & 12 & \checked & - & ? & 19 h & ? & 2D & 1 DoF & - & 1 & \checked \\ 
\addlinespace[3pt]
\cline{2-16}
\addlinespace[3pt]
& \multicolumn{2}{c|}{\textbf{\ac{DSC} (Ours)}} & \textbf{Germany, USA} & \textbf{I, C, P, H$^*$} & \textbf{5} & \textbf{\checked} & \textbf{\checked} & \textbf{\checked} & \textbf{15 h} & \textbf{175 k} & \textbf{3D} & \textbf{3 DoF} & \textbf{\textit{l, w, h}} & \textbf{14} & \textbf{\checked} \\ 
\bottomrule
\end{tabular}
\vspace{-5px}
\label{tab:datasets_overview}
\end{table*}

\section{The \acl{DSC}}
\subsection{Description}
\ac{DSC} counts about 15 hours of data across five different locations in Germany and the United States, with a total of 177,151 unique trajectories. The dataset covers diverse scene types including parking area (\acs{DSC-SIFI}), inner-city environment (\acs{DSC-MUC}), interesting non-signalized intersections (\acs{DSC-STR}, \acs{DSC-SFO}), and federal highway (\acs{DSC-BER}), capturing a wider range of scenarios compared to specialized datasets that focus on single environment types.

The \ac{DSC} dataset presents the largest classification framework of drone-based datasets including 14 separate categories. The main categories consist of pedestrians (140,227), bicycles (17,736), cars (13,241), scooters (1,475), motorcycles (1,054), animals (677), trucks (475), buses (191), and other (2,075). This detailed classification system allows researchers to perform in-depth analysis of interactions among various traffic participants.

A distinguishing feature of \ac{DSC} is its 3D data representation. Unlike existing datasets that typically provide 2D or 2.5D annotations, we deliver full 3D scene information including detailed mesh representations and precise 3D bounding boxes for all tracked objects.

The average duration of the trajectories is very heterogeneous between locations and object categories and goes up to 984 seconds. In total, 5,395 km of trajectory data are contained in this dataset. The motion data includes every tracked object with position, velocity, acceleration, and orientation, along with frame ID, track ID, and category.
\subsection{Accuracy}

We further validated the accuracy of our dataset by performing \ac{GPS} and 3D reconstruction error analyses on all locations. The \ac{RMSE} values, derived from \ac{BA} termination residuals, show a 1.928 meters deviation between reconstructed poses and consumer-grade GPS coordinates, and a 3D reconstruction error of 15 centimeters. It should be noted that there are techniques to further optimize the accuracy of the 3D reconstruction to centimeter precision, such as using \acp{GCP}, which are beyond the scope of this paper. \Cref{tab:datasets_reconstruction_accuracy} presents a summary of the \ac{RMSE} value of both metrics for each location.
\begin{table}[htbp]
\vspace{-5px}
\centering
\caption{GPS and 3D \ac{RMSE} values for different locations (in meters)}
\label{tab:datasets_reconstruction_accuracy}
\begin{tabular}{c|c|c|c|c|c|c}
\toprule
 & \textbf{SIFI} & \textbf{MUC} & \textbf{STR} & \textbf{BER} & \textbf{SFO} & \textbf{All} \\ \midrule
GPS RMSE & 1.76 & 2.069 & 2.421 & 0.964 & 2.427 & 1.928 \\ 
3D RMSE & 0.176 & 0.021 & 0.322 & 0.169 & 0.083 & 0.154 \\
\bottomrule
\end{tabular}
\vspace{-15px}
\end{table}

The detection validation was performed by comparing the positional median error between predicted detections and manually annotated detections of the same objects. Following an active learning strategy, our pipeline achieves a NuScenes score of 97\%~\cite{nuScenes}, evaluated in 3D space rather than pixel space as in~\cite{rwth_aachen}. Note that images of this dataset were included in the training of the detection network.

For qualitative validation, we provide two visualization tools. First, a web-based interface that displays 3D bounding boxes both on the 3D mesh and as projections onto video frames. Second, a visualization script that functions as a video stream player. These tools enable users to verify the spatial consistency and accuracy of our annotations across different modalities.
\subsection{Statistics}
This section provides a comparative analysis of the \ac{DSC} dataset against several publicly available trajectory datasets. As illustrated in \Cref{tab:datasets_overview}, \ac{DSC} is diverse in terms of countries and scene types, features a comparable dataset length to the largest set of tracks, includes the broadest variety of classes, and is the only dataset that offers 3D tracks. 

Further, we compare these datasets according to several key characteristics and point out the unique advantages offered by \ac{DSC}. For each of our dataset locations, we select a publicly available dataset of a similar scene type and compare it to ours.

\subsubsection{Trajectory Count and Class Distribution}
\begin{table}[htbp]
\vspace{-5px}
\centering
\caption{Number of trajectories per class for each dataset.}
\label{tab:trajectories}
\begin{tabular}{l|r|r|r|r|r}
\toprule
\multirow{2}{*}{\textbf{Class}} & \multirow{2}{*}{\makecell{\textbf{SIND} \\ \cite{SIND}}} & \multirow{2}{*}{\makecell{\textbf{AD4ACHE} \\ \cite{AD4CHE}}} & \multirow{2}{*}{\makecell{\textbf{OPENDD} \\ \cite{openDD}}} & \multirow{2}{*}{\makecell{\textbf{CITR / DUT} \\ \cite{CITR_DUT}}} & \multirow{2}{*}{\textbf{DSC3D}} \\ 
& & & & & 
\\ \midrule
Animal & 0 & 0 & 0 & 0 & \textbf{677} \\
Bicycle & 49 & 0 & 28 & 0 & \textbf{17,736} \\
Bus & 5 & \textbf{939} & 13 & 0 & 191 \\
Car & 1,647 & \textbf{42,516} & 1,013 & 95 & 13,241 \\
Motorcycle & 128 & 0 & 10 & 0 & \textbf{1,054} \\
Pedestrian & 142 & 0 & 30 & 2,111 & \textbf{140,227} \\
Scooter & 0 & 0 & 0 & 0 & \textbf{1,475} \\
Truck & 81 & \textbf{10,306} & 28 & 0 & 475 \\
Other & 0 & 0 & 121 & 0 & \textbf{2,075}\\
\bottomrule
\end{tabular}
\vspace{-15px}
\end{table}

\Cref{tab:trajectories} summarizes the number of trajectories per class for each dataset. Although we have more categories than the others, we standardize by merging the classes that belong to the same parent class. Our dataset significantly surpasses other datasets in capturing diverse and underrepresented classes, such as "Animal", "Scooter", and "Other" while maintaining a competitive number of trajectories for common traffic participants like "Car" and "Pedestrian".

\subsubsection{Trajectory Duration and Distance Distribution}
\begin{figure}[t]
    \centering
    \vspace{5px}
    \includegraphics[width=\linewidth]{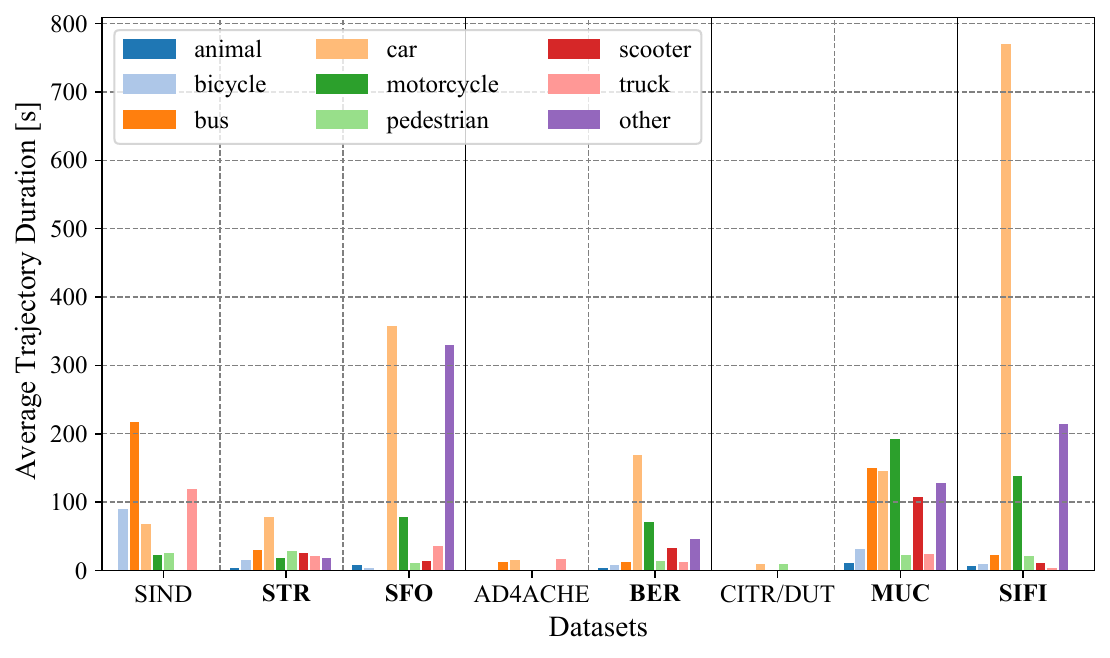}
    \vspace{-25px}
    \caption{Average of trajectory duration per class across datasets.}
    \vspace{-5px}
    \label{fig:trajectory_length}
\end{figure}

\Cref{fig:trajectory_length} compares the average trajectory durations across different datasets for several classes. Clearly, \ac{DSC} subsets like \ac{DSC-SIFI} and \ac{DSC-SFO} capture significantly longer trajectories for classes such as "Car" and "Motorcycle," pointing to their suitability towards the modeling of long-term movements with varied behaviors compared to other datasets. Furthermore, in \ac{DSC-MUC}, the duration of all classes is longer compared to the CITR/DUT~\cite{CITR_DUT} dataset, which highlights that objects are tracked over larger distances and for longer periods.

\begin{figure}[t]
\centering
\includegraphics[width=\linewidth]{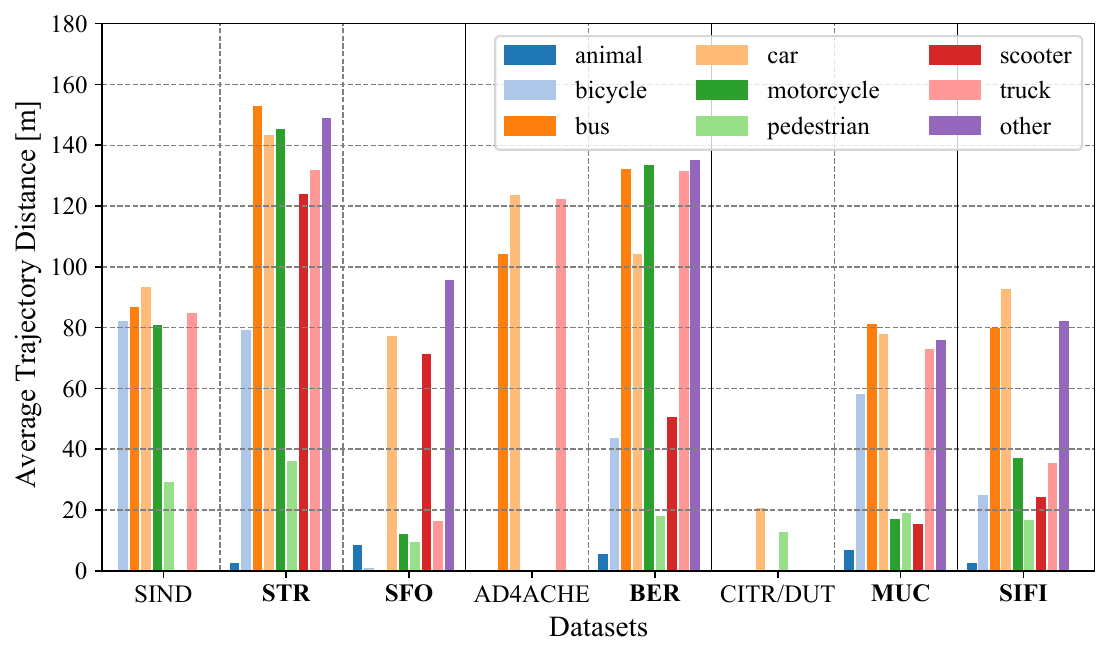}
\vspace{-25px}
\caption{Average trajectory distances per class across datasets.}
\vspace{-15px}
\label{fig:trajectory_distance}
\end{figure}

\begin{figure}[ht]
\centering
\vspace{5px}
\includegraphics[width=\linewidth]{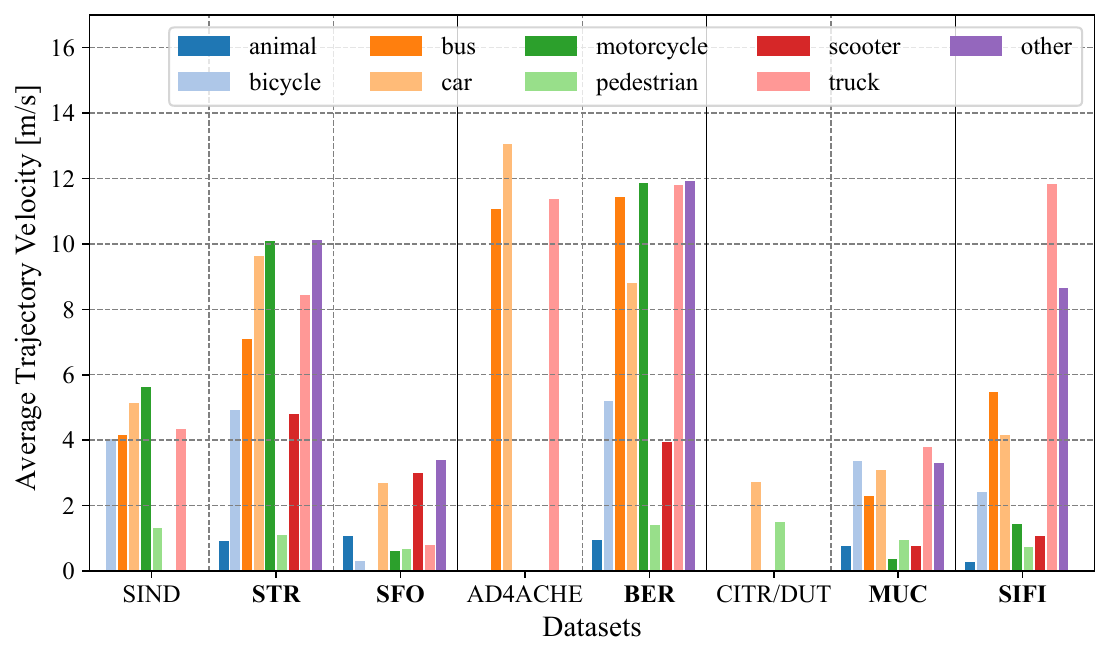}
\vspace{-25px}
\caption{Average velocity per class across datasets.}
\vspace{-9px}
\label{fig:avg_velocity}
\end{figure}

Averaged trajectory distances for all classes on several datasets are shown in \Cref{fig:trajectory_distance}. Notably, subsets of \ac{DSC} such as \ac{DSC-BER} and \ac{DSC-STR} have much longer trajectory distances for most classes, reflecting their ability to track objects over long spatial ranges. Also, the trajectory distances are always longer for \ac{DSC-MUC} in comparison with the CITR/DUT~\cite{CITR_DUT} dataset, which further justifies the effectiveness of our pipeline for tracking objects over larger areas.

\subsubsection{Average Velocity}
Averaged velocity for each class across datasets is shown in \Cref{fig:avg_velocity}. High variance of the velocities across all locations and classes visualizes the diversity of our dataset. Specifically, in \ac{DSC-SFO} the driving movements are slower because of steep roads and braking of vehicles when approaching the intersection. In the inner-city location \ac{DSC-MUC}, all trajectories are slower because of the high number of pedestrians. Road users were cautious and favored safety.

The given comparative analysis explains a number of essential advantages of \ac{DSC} dataset, as it stands out from other already existing trajectory datasets. These characteristics make \ac{DSC} a valuable resource for various applications.
\section{Applications}
\subsection{DeepUrban: A Benchmark for Planning and Prediction}
\ac{DSC} lays the foundation for the development of the DeepUrban \cite{deepurban} benchmark, designed to advance trajectory prediction and planning in complex traffic scenarios. The benchmark is created by extracting 20-second scenarios from \ac{DSC-MUC}, \ac{DSC-SIFI}, \ac{DSC-STR}, and \ac{DSC-SFO}. With \ac{DSC}’s extensive area coverage, nearly all vehicles in a scene can be treated as potential ego-vehicles, yielding a larger dataset than traditional perception datasets. Evaluation metrics such as \ac{ADE}, \ac{FDE}, and Collision Score were used to measure performance. In particular, it is shown in \cite{deepurban} that augmenting NuScenes \cite{nuScenes} training data with DeepUrban improved \ac{ADE} and \ac{FDE} scores by 44.1\% and 44.3\%, respectively, demonstrating \ac{DSC}’s utility for robust and generalizable benchmarks.
\subsection{Understanding Human Driving}
In a subsequent study, Kurenkov et al. \cite{safety_compliance} performed a comparative evaluation of human compliance with traffic and safety rules on several trajectory datasets like Argoverse~2~\cite{Argoverse2}, Lyft~\cite{lyft}, DeepUrban~\cite{deepurban} extracted from \ac{DSC}, and others. They have analyzed gap distance, \ac{TTC}, \ac{PET}, and adherence to German traffic regulations. The results showed that datasets like Argoverse~\cite{Argoverse2} and Lyft~\cite{lyft} have a lot of velocity and acceleration outliers while \ac{DSC} provides precise and realistic driving behaviors.

\subsection{Scenario Mining}
\begin{figure}[htbp]
    \centering
    \includegraphics[width=  \linewidth]{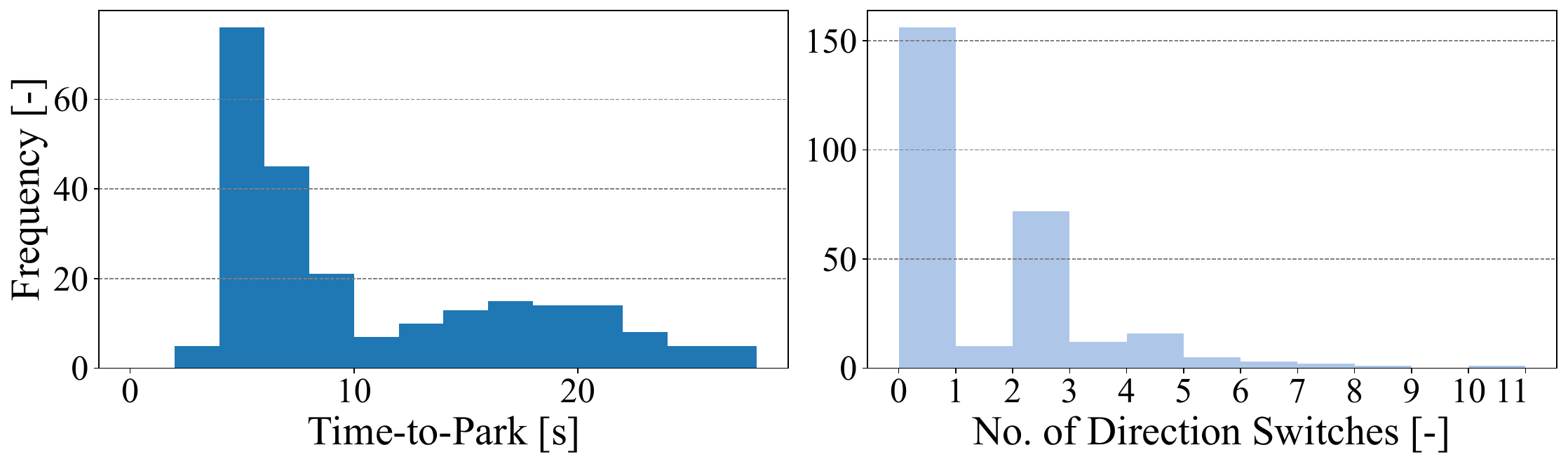}
    \vspace{-20px}
    \caption{Distribution of Time-to-Park (left) and Number of Direction Switches (right) of parking scenarios mined on \ac{DSC-SIFI}.}
    \label{fig:parking_stats}
\end{figure}

\begin{figure}[htbp]
    \centering
    \vspace{5px}
    \includegraphics[width=  \linewidth]{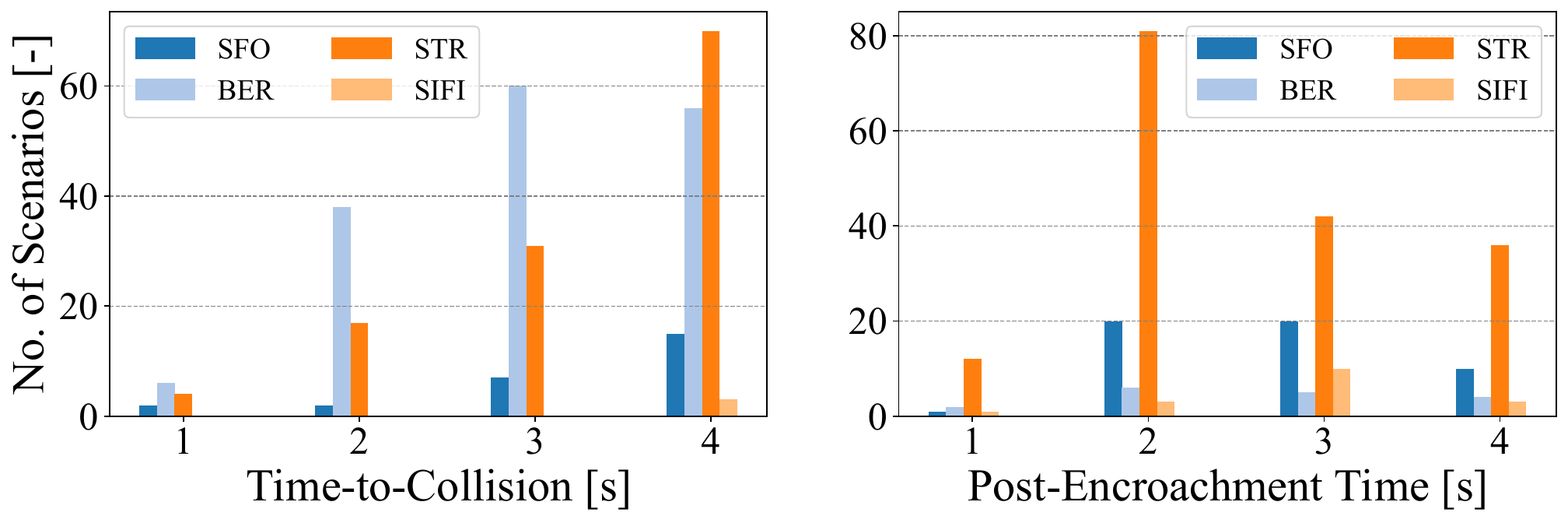}
    \vspace{-20px}
    \caption{Distribution of \ac{TTC} and  \ac{PET} of critical scenarios extracted from the dataset.}
    \vspace{-15px}
    \label{fig:ttc_pet}
\end{figure}

Figures~\ref{fig:parking_stats} and~\ref{fig:ttc_pet} provide the analysis of parking scenarios and critical events in our dataset. The distribution of parking time in \ac{DSC-SIFI} showcases that most maneuvers are completed within 10 seconds, whereas few exceed 30 seconds. The number of direction switches while parking is mostly one or two, reflecting typical low-risk parking behavior. These human demonstrations of solving challenging parking situations can help improve the performance of current automated parking systems, which often struggle with geometric analysis.

Concerning critical events, \Cref{fig:ttc_pet} presents the distributions of \ac{TTC} and \ac{PET}. \ac{DSC-STR} scenarios are predominant in the critical \ac{TTC} range of 2-4 seconds and constantly show close proximity encounters in \ac{PET} of 1-2 seconds. However, \ac{DSC-SIFI} scenarios present higher values of \ac{PET}, confirming their nature.

\begin{figure}[htbp]
    \centering
    \includegraphics[width=  \linewidth ]{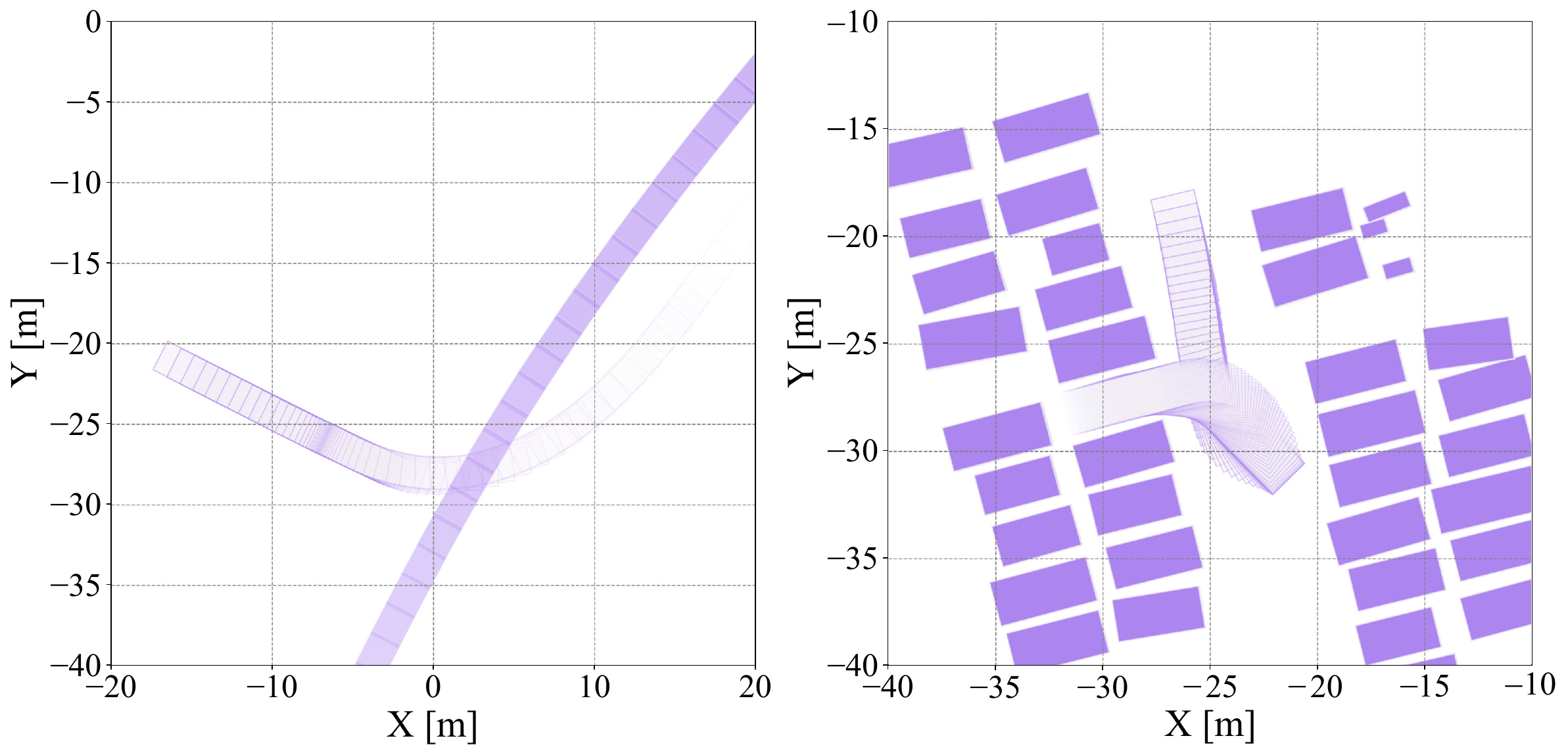}
    \vspace{-20px}
    \caption{Critical scenario with \ac{PET}=1.1s mined on \ac{DSC-STR} (left) and parking scenario with two direction switches mined on \ac{DSC-SIFI} (right).}
    \vspace{-10px}
    \label{fig:scenario_miner}
\end{figure}

Figure~\ref{fig:scenario_miner} exemplifies these results by showing representative examples for both a critical scenario at \ac{DSC-STR} with low \ac{PET} and a typical parking maneuver with direction switches at \ac{DSC-SIFI}. These distributions and examples therefore suggest that our dataset covers both routine parking maneuvering and safety-critical intersection scenarios and is thus suitable for a wide range of autonomous driving tasks.

\subsection{Generative Reactive Traffic Agents}
\begin{figure}[htbp]
    \centering
    \vspace{5px}
    \includegraphics[width= 0.49\linewidth]{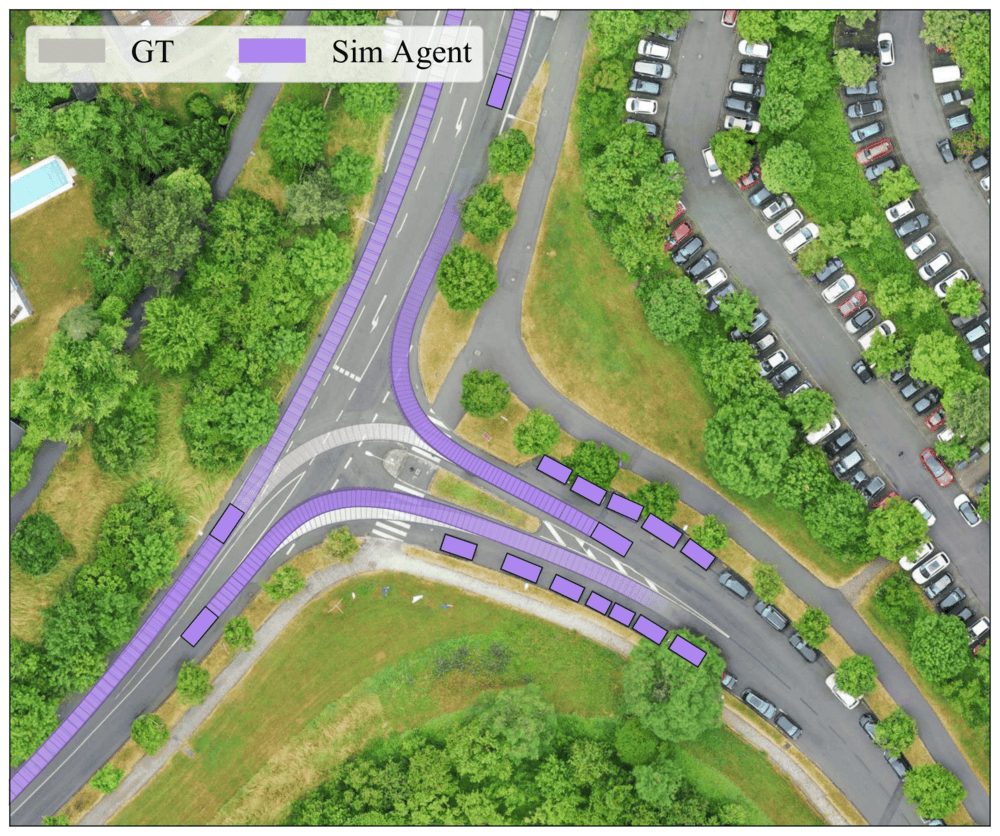}
    \includegraphics[width=  0.49\linewidth]{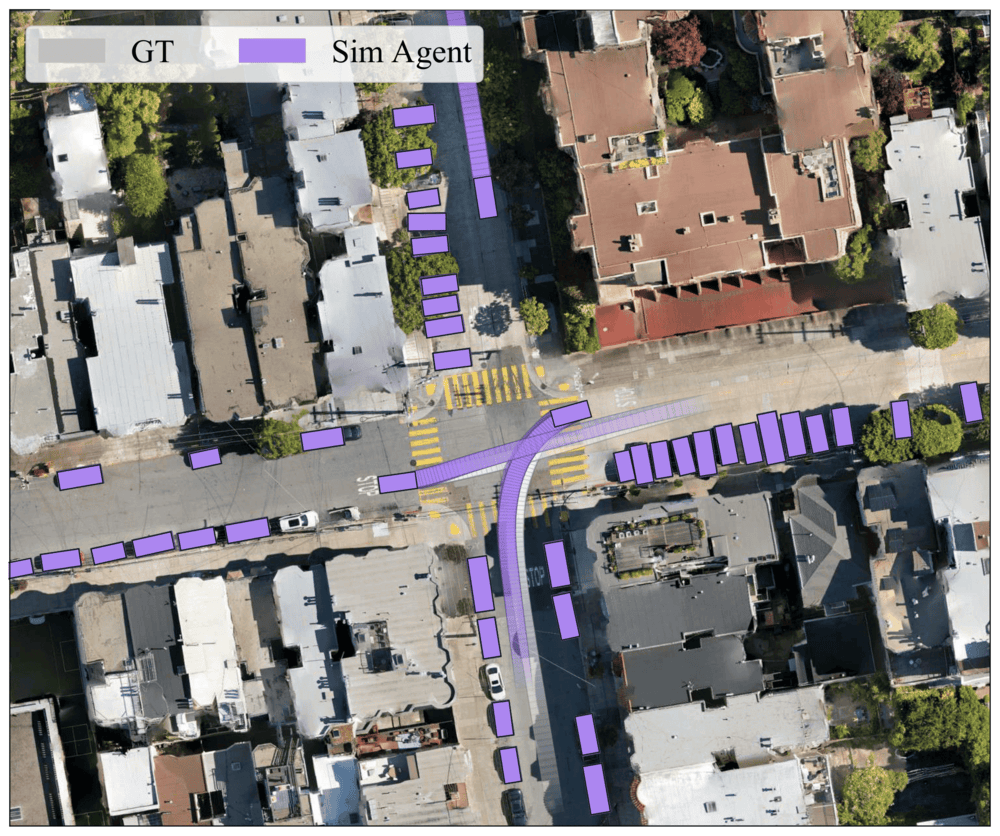}
    \vspace{-20px}
    \caption{Synthetic scenarios generated by traffic agents on \ac{DSC-STR} (left) and on \ac{DSC-SFO} (right).}
    \vspace{-15px}
    \label{fig:traffic_agents}
\end{figure}

One notable application of the dataset described above is the training of data-driven, reactive traffic agents. These models can be deployed in closed-loop simulation environments where agents dynamically interact with one another and their surroundings. This approach captures critical features of real-world traffic systems, addressing the lack of realism often seen in traditional rule-based simulations. State-of-the-art models, such as those proposed in \cite{zhou2024behaviorgptsmartagentsimulation}, \cite{huang2024versatilebehaviordiffusiongeneralized}, and \cite{zhang2024trafficbotsv15trafficsimulation}, can be trained on these datasets to emulate realistic human driving behavior, subsequently generating diverse and interactive traffic scenarios.
Figure \ref{fig:traffic_agents} illustrates examples of synthetically generated scenarios created by agents trained on the \ac{DSC-STR} and \ac{DSC-SFO} datasets using DeepScenario's proprietary generative traffic agent model. The model uses the previous path of every agent as an initial seed to predict their future movement. It simulates how agents will react to and learn from the actions of nearby agents, generating realistic trajectories with coherence in the dynamics of the scene, though not necessarily similar to the originally captured trajectories. 

\section{Conclusion}
In this work, we presented, for the first time, a 3D trajectory dataset captured from the drone perspective called \ac{DSC}, which presents outstanding diversity in various large sets of environments, such as urban intersections, inner-city scenes, parking areas, and federal highways from both Germany and the United States. The diverse dataset comprises more than 175,000 distinct trajectories further categorized into 14 categories and features a variety of terrains at different elevations. We developed a high-precision monocular tracking pipeline that achieves remarkable positional accuracy in 3D space. Our experiments demonstrated the potential of the dataset across a wide range of applications, including scenario mining, trajectory prediction, traffic simulation, and training generative models of traffic agents that can produce realistic behaviors across diverse contexts.
For future work, we plan to integrate traffic signaling information from signalized intersections and further improve our detection and tracking pipeline through advanced deep learning techniques.

\bibliographystyle{IEEEtran}
\bibliography{references} 

\end{document}